\documentclass[11pt]{article}
\usepackage{mymacros}
\title{Opening the Black Box: predicting the trainability of deep neural networks with reconstruction entropy}

\author[1]{Yanick Thurn,}
\affiliation[1]{Institute for Theoretical Physics and Astrophysics and \\ Würzburg-Dresden Cluster of Excellence ct.qmat, Julius-Maximilians-University W\"urzburg, Am Hubland, 97074 W\"urzburg, Germany}

\author[2]{Ro Jefferson,}
\affiliation[2]{Institute for Theoretical Physics, and Department of Information and Computing Sciences,\\Utrecht University, Princetonplein 5, 3584 CC Utrecht, The Netherlands}

\author[1]{and Johanna Erdmenger}

\abstract{An important challenge in machine learning is to predict the initial conditions under which a given neural network will be trainable. We present a  method for predicting the trainable regime in parameter space for deep feedforward neural networks (DNNs) based on reconstructing the input from subsequent activation layers via a cascade of single-layer auxiliary networks. We show that a single epoch of training of the shallow cascade networks is sufficient to predict the trainability of the deep feedforward network on a range of datasets (MNIST, CIFAR10, FashionMNIST, and white noise), thereby providing a significant reduction in overall training time. We achieve this by computing the relative entropy between reconstructed images and the original inputs, and show that this probe of information loss is sensitive to the phase behaviour of the network. We further demonstrate that this method generalizes to residual neural networks (ResNets) and convolutional neural networks (CNNs). Moreover, our method illustrates the network's decision making process by displaying the changes performed on the input data at each layer, which we demonstrate for both a DNN trained on MNIST and the vgg16 CNN trained on the ImageNet dataset. Our results provide a technique for significantly accelerating the training of large neural networks. 
}

\begin{document}

\maketitle


\section{Introduction}

Despite the impressive empirical success of deep neural networks (DNNs), the theoretical principles underlying deep learning remain relatively poorly understood. A particularly important challenge for machine learning is to determine for which initial configurations neural networks are optimally trainable. Current state-of-the-art approaches aim at optimizing these configurations by grid searches over parameter space that involve training multiple network configurations (see e.g.~\cite{optuna_2019}). These grid searches may however represent an obstacle to the cost effectiveness of neural networks, and are not even feasible for larger networks used in industry. To reduce this computational burden, significant theoretical effort has been devoted to determine the initial conditions under which a given network will be trainable. Here, we present an alternative approach that significantly accelerates the process of selecting suitable parameters by predicting trainable regions in phase space \emph{without any training} of the DNN.

\textbf{Related work.} The vast majority of existing theoretical and experimental approaches take place in the infinite-width approximation, in which the number of neurons $N$ per layer is taken to infinity.\footnote{This framework also holds if one allows different numbers of neurons $N_\ell$ in different layers $\ell$, provided all $N_\ell\to\infty$ at the same rate. Informally, all one requires is that the distribution of each layer is approximately Gaussian.} Due to the central limit theorem, the distribution of neurons in each layer becomes normal as $N\to\infty$, and hence the network becomes a Gaussian process (see \cite{rasmussen2005gaussian} for a comprehensive treatment). This mathematical simplification allows many results to be derived exactly. 

One important line of work in this direction centers on the neural tangent kernel (NTK) \cite{Jacot2018}. In the infinite-width limit, the NTK becomes frozen, obtaining a duality between training the network under gradient descent and kernel methods, allowing one to study the training dynamics analytically in this regime, see e.g. \cite{Nakazato2024AST,Zhao2021ZerOII}. However, results based on the constancy of the NTK become less accurate if the ratio of depth to width is finite, as is the case in real-world networks. For example, of particular relevance to the present work is \cite{Seleznova2022}, which examines how the behaviour of the NTK changes in different regions of phase space (we shall elaborate on this momentarily) when \emph{both} width and depth are taken to infinity. More generally however, networks in the infinite-width NTK parametrization do not actually learn features \cite{Yang2020FeatureLI}, which further underscores the need for alternative approaches.

Another important approach to the problem of fine-tuning large (industry-scale) networks is hyperparameter transfer: the basic idea is to find the optimum initialization for a small neural network, and then deduce how that parametrization will scale to a large one. This approach has been spearheaded by Yang and collaborators \cite{Yang2022TensorPV} (see also\cite{Mei2018AMF,Chizat2018OnTG} for previous work), who demonstrated that so-called maximal-update parametrization of shallow networks can be successfully transfered to larger models. This has since been generalized to deeper networks in \cite{Yang2023TensorPV,Bordelon2023DepthwiseHT}. Notably however, as commented in sec. 1.3 of \cite{Yang2022TensorPV}, this approach breaks down when the block depth of the ResNet is greater than 1. Additionally, as will be seen below, this approach is still much more computationally intensive than the one we present here, since one still has to find the optimal conditions for, and successfully train, the smaller network; for the case studied in \cite{Yang2022TensorPV} for example, this required 5000 training steps, whereas our approach allows one to determine this in a single epoch.

A particularly relevant application of trainability prediction is to signal propagation in ResNets. For example, \cite{Yang2017MeanFR} showed that the optimal intialization variances for ResNets may depend on the depth, which poses a challenge for existing theoretical approaches (cf. \cite{Yang2022TensorPV} above). Related work \cite{Li2021TheFI,Li2022TheNC} seeks to make progress by again taking both the width $N$ and depth $L$ to infinity, so that the parameter $L/N$ remains finite, as mentioned above. Interestingly, these limits appear to commute in the special case of ResNets \cite{Hayou2023WidthAD}. In general, this is not the case, since the non-linearities accumulate with depth, causing a loss of theoretical control. At a technical level, this is because the finite-width corrections to the Gaussian Process approximation can be written as a parametric expansion in $L/N$ \cite{Roberts:2021fes,Grosvenor:2021eol}. Nonetheless, while idealized, this limit remains an important tool for studying these networks due to the afforded level of theoretical control.\\

To summarize these developments, it is highly desirable -- and even necessary, in the case of larger networks used in industry -- to find the optimal initial network configuration prior to performing training. Here, we present a highly efficient alternative to the approaches discussed above. In our experiments for the MNIST and CIFAR10 datasets for example, we find a speed-up of up to two orders of magnitude.  Moreover, our approach allows to identify which features of the data stored in the hidden layers are particularly important for the network's decision-making process. While this work is primarily a proof of concept, we have also generalized our method to CNNs and ResNets, and tested it on the additional datasets FashionMNIST and ImageNet.

Our approach is based on a physics-inspired concept that we name \emph{reconstruction entropy}, and builds on the rapidly growing intersection of physics and machine learning in the past few years, wherein concepts and ideas from physics have proven increasingly useful for improving both our theoretical understanding and the practical use of DNNs. The relationship between neural networks and statistical mechanics in particular has a rich history (see \cite{Bahri-review} for a review, as well as references therein), and connections between deep learning and the renormalization group (RG) have been explored in, among others, \cite{mehta2014exact,Koch-Janus_2017,LenggenhagerOT,Erdmenger:2021sot,Kline_2022,Berman:2022uov,Howard:2024bbi}. More recently, various correspondences between DNNs and quantum field theory have been developed (see \cite{Halverson:2020trp,Grosvenor:2021eol} and subsequent works), providing a rigorous explanatory framework for the finite-width effects discussed above, and techniques from QFT have been applied more generally at both a structural \cite{Roberts:2021fes,Berman:2024pax,Berman:2022mak} and dynamical \cite{Cohen,Howard:2024bbi} level.

More specifically, the starting point for the present work is the observation that, as alluded above, many deep neural networks exhibit a phase transition between an ordered and chaotic phase \cite{poole2016exponential,schoenholz2017deep,xiao2018dynamical,chen2018dynamical}. These are characterized by vanishing and exploding gradients, respectively, and hence networks initialized in either phase are difficult or impossible to train. Intuitively, this is because information about the input data cannot propagate through the network to affect the cost function, since correlations either die-out in the ordered (low-temperature) phase, or are washed-out by noise in the chaotic (high-temperature) phase. However, the boundary between these phases is described by a continuous phase transition at which the correlation length diverges, allowing theoretically infinite signal propagation, and hence trainability of arbitrarily deep networks. In other words, a precondition for trainability is that the depth of the network must not exceed the scale set by the correlation length. Thus, increasingly deep networks must be initialized closer to this so-called ``edge of chaos'' in order to be trainable. Determining the location of the edge of chaos in phase space is thus of significant practical benefit, since it avoids the need for computationally costly grid searches over the space of parameters.\footnote{Here we mean the weights and biases that parametrize the phase space. The selection of appropriate hyperparameters such as batch size, learning rate, etc., are of secondary importance in the sense that no arrangement of hyperparameters can redeem a network in the untrainable regime. Conversely, if the network does not appear to train at some arbitrary point in phase space, it is \emph{a priori} unclear whether the selection of parameters or hyperparameters is the problem. The present work aims to resolve this. For related work on trainability vs. generality, see e.g. \cite{Xiao2020} and references therein. \label{ft:hyper}}

Initial works \cite{poole2016exponential,schoenholz2017deep} on deep feedforward networks determined the location of the critical line in above-mentioned idealized regime in which the width of the network is taken to infinity. However, as real-world networks are necessarily of finite width, the theoretical prediction for the critical point as well as the correlation length itself differs from empirical results; this can be seen already in the initial works above, and was shown even more starkly in \cite{Erdmenger:2021sot}. While significant mathematical machinery has been brought to bear on the problem of predicting finite-width corrections to this central limit theorem\footnote{Most works in the machine learning literature refer to the infinite width limit as mean field theory. As discussed in \cite{Grosvenor:2021eol} however, this is not quite right for two reasons: conceptually, there is no continuum limit and hence no \emph{bona fide} fields in these previous works; and technically, the mean field theory is also not necessarily the same as the result from the central limit theorem. Since the latter is ultimately all that is required, we refer to this line of work exclusively under this banner.} result \cite{Grosvenor:2021eol,Naveh_2021,bordelon2023dynamics, PhysRevLett.120.248301, Li_2020}, accurately determining the trainable regime for a given network architecture remains an open challenge.

In light of the above difficulties, it is pertinent to ask whether other tools from physics may be applied to determine the location of the edge of chaos prior to training (that is, sparing the need for computationally costly grid searches over the entire phase space). Ultimately, the problem is to determine the conditions under which information, suitably quantified, can propagate through the entire network. In \cite{Erdmenger:2021sot}, parallels between subsequent layers in a DNN and successive steps of real-space RG were found by analyzing the information flow through the network via the Kullback-Leibler (KL) divergence or relative entropy,
\begin{equation}
	D\left(p(x)||q(x)\right)\coloneqq\sum_{x\in\chi}p(x)\ln\frac{p(x)}{q(x)}~,
	\label{eq:KL}
\end{equation}
where $p(x)$ and $q(x)$ are probability mass functions over the sample space $\chi$. For distributions normalized with respect to the same measure, the KL divergence is non-negative, and zero if and only if $p=q$, Interestingly, while the results in \cite{Erdmenger:2021sot} for both DNNs and RG appear
qualitatively identical to each other, the relative entropy appears insensitive to the location in phase space, and therefore could not be used to predict the location of the critical line. However, that work operated in the infinite width approximation in order to obtain analytic expressions for the relative entropy between layers, which were then implemented in code. An explanation for the lack of sensitivity to the phase transition may lie in the fact that \cite{Erdmenger:2021sot} considers the activations in each layer to be Gaussian. As a Gaussian is characterized by a single cumulant, this is not sufficient for storing higher-order correlations. 

In the present work, we present a new approach, referred to as \textit{reconstruction entropy}, that involves working  directly with the (suitably normalized) neuron activations themselves as comprising a probability mass function - i.e., a discrete probability distribution. This naturally includes higher-order correlations. Moreover, this approach overcomes the issue that the entropy, directly calculated between different layer activations, results in a spurious increase in the relative entropy that does not reflect meaningful information loss.
As a trivial example for this issue, consider the vector $\mathbf{v}=(8,5,12,12,15)^\mathrm{T}$ encoding the word ``HELLO'' encoded with the alphabet, and a function that returns $f(\mathbf{v})=(72,69,76,76,79)^\mathrm{T}$, i.e.~encoding in ascii. Intuitively, it seems that the information content in both vectors is the same, but the Kullback-Leibler divergence \eqref{eq:KL} with $p(x)=\mathbf{v}/|\mathbf{v}|$ and $q(x)=f(\mathbf{v})/|f(\mathbf{v})|$ is approximately $0.046$. As a slightly more realistic example, shifting an MNIST digit 1 pixel to the left between layers would result in a finite KL divergence, but a simple translation of the input image does not meaningfully alter what we intend for the network to learn.\footnote{See \cite{Cheng:2019xrt} for a physics-inspired approach to equivariance in neural networks; here, the problem is more generally that the information we wish to capture may be stored differently in different layers, not necessarily related by any convolution.}

\begin{figure}[H]
    \centering
    \includegraphics[width=\linewidth]{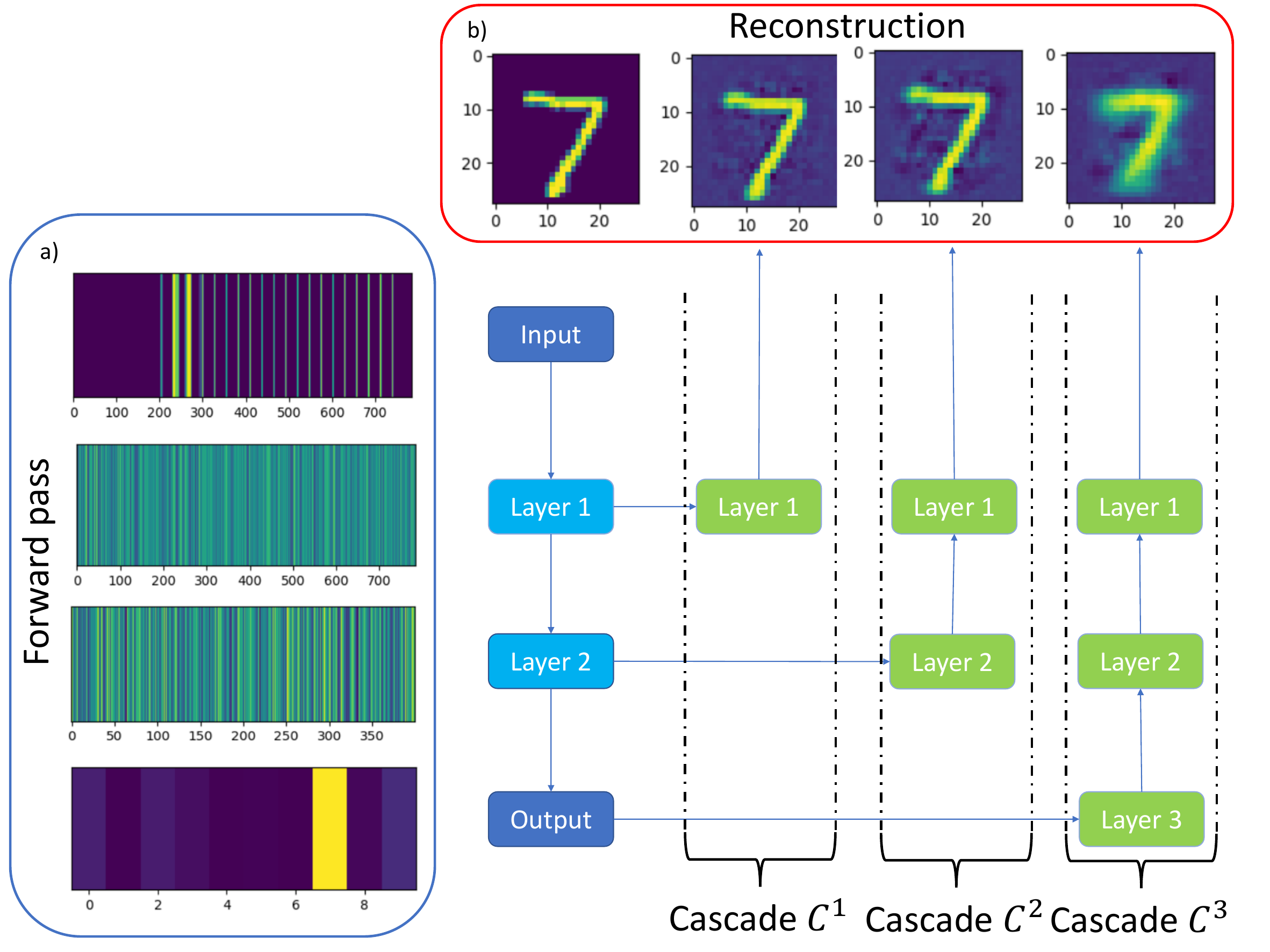}
    \caption{\textbf{Illustration of the reconstruction process} for a feedforward network trained on MNIST data. In this example, an image of the digit 7 is fed into the DNN (blue column), consisting of two hidden layers with widths 784 ($\ell\!=\!1$) and 400 ($\ell\!=\!2$). The blue box on the far left illustrates the vectorized activations of the neurons at each layer, including the input (data) and output (one-hot vector). The auxiliary cascade networks are shown in green, with their corresponding reconstructions in the red box at the top (the left-most image therein is the original input image for reference). For each layer $\ell$ of the DNN, we have an $\ell$-layer cascade network that reconstructs the input image from the activations $\textbf{z}^\ell$. The increasingly blurred reconstruction images reflects loss of information in feedforward pass. We emphasize that the individual layers of the cascades are trained separately, such that layer $\ell$ of a given cascade is trained to reconstruct $\mathbf{z}^{\ell-1}$ from $\mathbf{z}^\ell$. \label{fig:cascade-example}}
\end{figure}

The reconstruction entropy (RE) approach  that successfully addresses this issue involves computing the relative entropy between the input to a given layer and a reconstruction thereof by auxiliary networks that we refer to as \emph{cascade networks}; see fig.~\ref{fig:cascade-example}. The basic idea of the reconstruction entropy approach is as follows: suppose we have an $L$-layer feedforward network, with vectorized two-dimensional images fed into layer $\ell\!=\!0$. The neurons, i.e., the activations, in layer $\ell$ are given by
\begin{equation}
	\mathbf{z}^\ell\coloneqq\phi\left(\mathbf{W}^\ell\cdot\mathbf{z}^{\ell-1}+\mathbf{b}^\ell\right)~,
	\label{eq:DNN}
\end{equation}
for $\ell\in[1,L]$, where $\mathbf{z}^\ell\in\mathbb{R}^N$ is the vector of activations, $\mathbf{z}^{\ell-1}\in\mathbb{R}^M$ are the preactivations (i.e., the activations of the previous layer), $\mathbf{W}$ is an $N\times M$ matrix of weights (connection strengths), $\mathbf{b}$ is an $N$-dimensional vector of biases (firing susceptibilities), and $\phi$ is a (generally non-linear) activation function. For concreteness, we will limit ourselves in this article to $\tanh(x)$, in keeping with previous theoretical works in this vein \cite{poole2016exponential,schoenholz2017deep,Erdmenger:2021sot,Grosvenor:2021eol} (see also \cite{Roberts:2021fes} for theoretical background for this choice in the present context). We will work in the setting of so-called random networks in which the weights and biases are initialized from zero-mean Gaussian distributions with variances $\sigma_w^2$ and $\sigma_b^2$, respectively, again keeping with these previous works.\footnote{For wide networks, these distributions converge to Gaussians within very few layers, so this assumption does not result in any loss of generality; see \cite{Erdmenger:2021sot,Bukva:2023ksv} for related discussions.}

 The activations of each layer $\ell$ are then fed as inputs to a shallow auxiliary network, which is trained to reconstruct the original preactivations $\mathbf{z}^{\ell-1}$. Note that the training of the auxiliary networks is done sequentially: that is, at $\ell\!=\!1$, a shallow cascade network $\mathcal{C}^1$ is trained to reconstruct the original input $\mathbf{z}^0=\mathbf{x}$ from the activations $\mathbf{z}^1$. At $\ell\!=\!2$, a second shallow network is trained to reconstruct $\mathbf{z}^1$ from $\mathbf{z}^2$. Once trained, this layer is combined with the previous auxiliary layer to form cascade $\mathcal{C}^2$ (see fig.~\ref{fig:cascade-example}), so that a reconstructed image can then be obtained from $\mathbf{z}^2$, and so on. Thus, the auxiliary cascade network $\mathcal{C}^\ell$ consists of $\ell$ individually trained layers, each of which acts as a shallow decoder for the corresponding layer in the feedforward network (henceforth referred to as ``the DNN''). In this sense, the relationship between the feedforward network and the cascades resembles that of an autoencoder, though in this case the activations of each individual layer in the DNN are taken as a sequence of latent spaces. Importantly, since the cascades are trained as a collection of shallow networks, training is extremely quick, and can be completed with high accuracy within a single epoch. Since the components (layers) of the cascades are trained individually, they are always in the trainable regime for any reasonable choice of parameter values.\footnote{It is possible for networks to be untrainable even at criticality if the activation function saturates \cite{Bukva:2023ksv}.} Furthermore, the relative entropy computed between the reconstructed image from cascade $\mathcal{C}^\ell$ and the original input provides a more indicative measure of meaningful information loss, as this minimizes artifacts associated with qualitatively identical content being encoded in different ways.\footnote{We note that computing the entropy without using the cascades, i.e., with the layers of the feedforward network itself, merely results in noise.} This is illustrated in fig.~\ref{fig:cascade-example} by the images reconstructed from deeper and deeper layers of the DNN becoming increasingly blurred as information is lost that the cascades are unable to reconstruct. To again employ an analogy with autoencoders, we do not want to compute the relative entropy between the data and the latent representation, but between the data and its reconstruction. We note that this method also avoids the constraint, previously imposed in \cite{Erdmenger:2021sot}, that the widths of all layers of the feedforward network be identical to avoid issues with normalizing the distributions, since here we are always computing the relative entropy between vectors in the same space. In what follows, when using the term  `relative entropy', we always refer to the relative entropy between the reconstruction and the input.\footnote{As an aside, we note that the Kullback-Leibler divergence has also been employed to improve trainability in \cite{Peer}, though this approach contrasts strongly with ours. There, a new, entropy-inspired term is added to the loss function, which contains another hyperparameter that must be obtained via an additional grid search. In contrast, we do not alter the loss function; furthermore, as emphasized below, the novel auxiliary cascades introduced in our approach significantly \emph{reduce} the need for such hyperparameter searches.}

\begin{figure}[H]
\thispagestyle{empty} 
    \centering
    \includegraphics[width=0.8\textwidth]{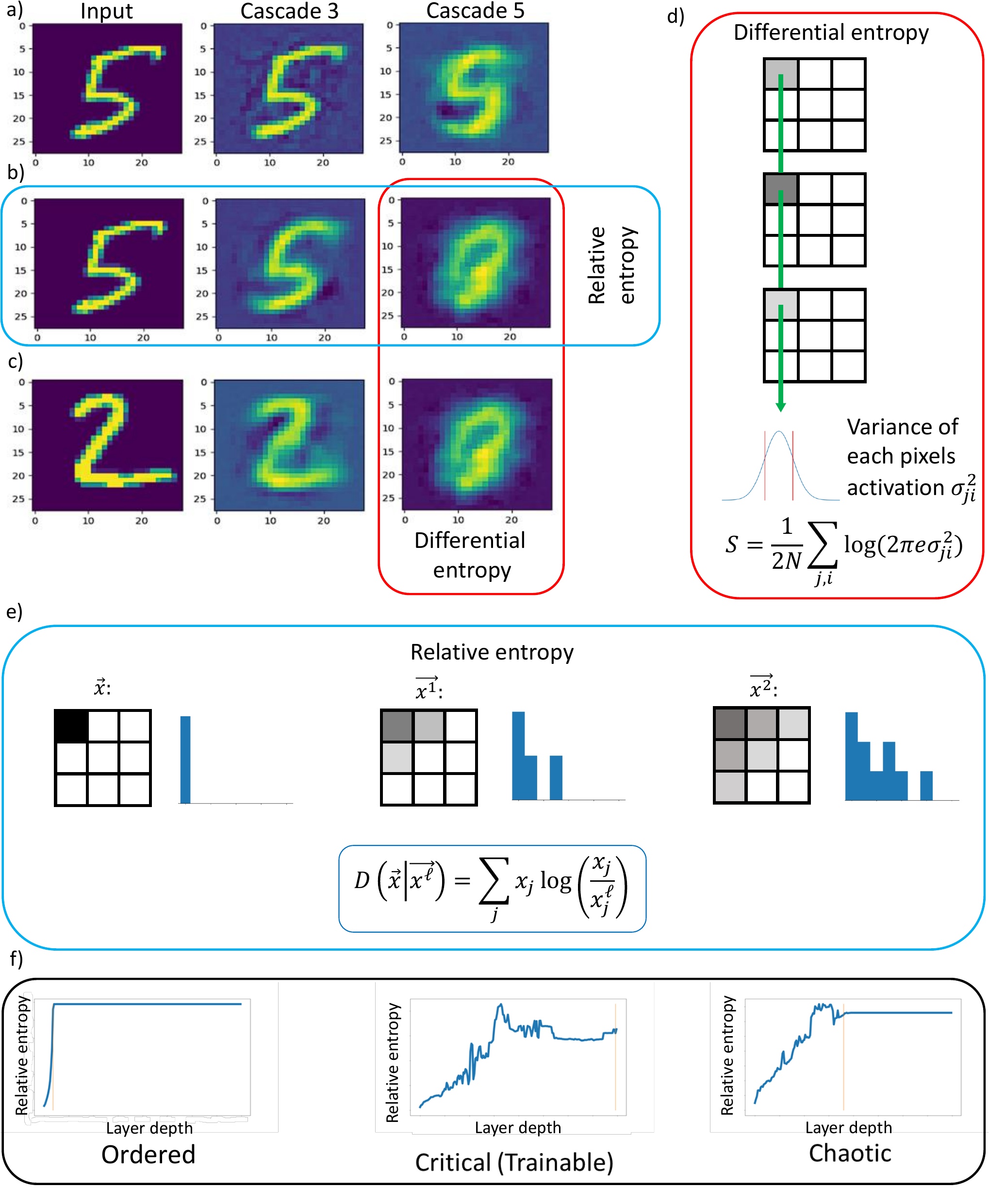}
    \caption{\textbf{Illustration of the two methods for computing the reconstruction entropy.} Panel (a) shows the reconstructed image from layers $\ell=0,3,5$ (left to right) for a trainable but untrained DNN. Panels (b) and (c) show the same for a network in the untrainable regime with two different inputs. One sees in the third column ($\ell\!=\!5$) that for untrainable networks, different inputs converge to a common reconstruction as information is lost. The red and blue boxes around images in panels (b) and (c) correspond to the computation of the differential and relative entropy, respectively, which are illustrated in more detail in panels (d) and (e). For the differential entropy, we construct a probability distribution function for each individual pixel, drawn from the set of reconstruction images. (Panel (d) illustrates this for a sample size of 3; in practice, we use a sample size of 100.)	For the relative entropy, we compare sequential reconstruction images with the corresponding input; note that the variation between individual pixels is also important in this approach, but as only two images are compared at a time, the probability mass function is discrete. The normalized activation
    for each pixel is
    represented by histograms in panel (e)). In the bottom panel (f), we show the relative entropy for an untrained network in the stable, critical, and chaotic regimes. The stable and chaotic regimes are characterized by saturation of the relative entropy after a certain point we call the cutoff. For networks at criticality, there is no such cutoff (by default, it is equal to the total depth $L$). This provides the basis for our prediction of trainability without needing to train the feedforward network; we only need to train the series of shallow networks (i.e., the individual layers) that comprise the cascades, cf.~fig.~\ref{fig:cascade-example}.
    \label{fig:cutoff_detection}}
\end{figure}

Regarding the trainability, the key observation is that the relative entropy, computed in this manner, exhibits qualitatively different behaviour depending on where the network is initialized in the phase space parametrized by $\left(\sigma_w^2,\sigma_b^2\right)$. This is illustrated in the last panel of fig.~\ref{fig:cutoff_detection}. In the stable regime, the KL divergence is characterized by a relatively sharp increase, and then remains unchanged for all subsequent layers. This is due to the rapid decay of correlations in the ordered phase: intuitively, fluctuations in the system are frozen in the low-temperature limit. Conversely, the chaotic phase is characterized by a more gradual but highly non-monotonic increase, reflecting the relatively large fluctuations that occur at high-temperature. Again however, the KL divergence eventually saturates at some finite (meaning, $\ell^* <L$) layer before remaining constant in all subsequent layers, as all information about the input has been washed out by noise. The point at which the KL divergence reaches this saturation point is called the \emph{cutoff}. For networks initialized near criticality however, the KL divergence never saturates, so there is no such cutoff. More precisely, since the correlation length will not be infinite in a real-world network due to finite-width effects, the cutoff exceeds the depth of the network. That is, our central claim is that the cutoff at which the relative entropy saturates provides a measure of the correlation length in the network, and hence a finite cutoff $\ell^*<L$ indicates that the network is not trainable; see fig.~\ref{fig:combination1}. We emphasize again that it is not necessary to train the feedforward network in order to observe this qualitative distinction between phases: rather, it is set by the location of the network in phase space, and can be extracted by instead training only the shallow components of the cascades.\footnote{Note that while this implies that a 50-layer DNN would need $50\times 1$ epochs of training (1 for each individual layer of the cascades), the training of the cascade layers can be done in parallel, and since each layer is treated as a shallow network, an epoch of training is orders of magnitude faster than an epoch of training on the feedforward DNN. We comment on further on the computational advantages in the Results below.}

However, while the KL divergence provides a clearer conceptual basis for this approach, in practice it is subject to some numerical instability.\footnote{The reason for this is that the reconstructed image to which all inputs converge after information is lost may happen to look more similar to certain input images than others. In fig.~\ref{fig:cutoff_detection} for example, the right-most reconstructed image in panels (b) and (c) has a higher KL divergence relative to a 1 than to an 8 or a 0. Thus, averaging over a large number of inputs does not solve this problem, since similarity to even a single class of digit is enough to yield fluctuations on the order of 10\%.} Accordingly, we have found that an alternative method for computing reconstruction entropy yields more consistent results. The basis for this is the differential entropy
\begin{equation}
S\left(\rho(x)\right)\coloneqq-\int_\chi\!\mathrm{d}x\,\rho(x)\ln \rho(x)~.
	\label{eq:diffent}
\end{equation}
Note that we have used $\rho(x)$ to denote the (continuous) probability distribution function appearing in this expression, to avoid confusion with the (discrete) probability mass function $p(x)$ appearing in \eqref{eq:KL}. We construct $\rho(x)$ as follows (see fig.~\ref{fig:cutoff_detection}).  For fixed $\ell$, we consider a set of reconstructed images obtained for a random selection of inputs. For all images in this set, choose the same pixel, denoted by its location $(i,j)$ (with $i,j\in\{1,\ldots,\sqrt{N}\}$ where, e.g., for MNIST $N\!=\!784$), and normalize so that $\rho_{ij}(x)$ is the distribution of values taken by pixel $(i,j)$ over the set of reconstructed images. If this set is sufficiently large, then $\rho$ becomes approximately Gaussian\footnote{We emphasize that we are \emph{not} referring to the distribution of pixels within a given image such as discussed in, e.g., \cite{Hyvrinen2009NaturalIS}.}, and therefore the entropy \eqref{eq:diffent} is simply
\begin{equation}
	S_{ij}\coloneqq S(\rho_{ij})=\frac{1}{2}\ln\left(2\pi e\sigma_{ij}^2\right)~,
	\label{eq:diffent2}
\end{equation}
where $\sigma_{ij}$ is the standard deviation of $\rho_{ij}(x)$. Empirical support for this statement is given in appendix \ref{sec:toldyouso}. We define the reconstruction differential entropy as the sum of the differential entropies for all pixels in the set of reconstructed images, $S\coloneqq\frac{1}{N}\sum_{ij}S_{ij}$. For a random collection of inputs, differences between pixel values between images will result in a relatively high variance. As shown in fig.~\ref{fig:cutoff_detection} however, as the information is lost in successive layers, the reconstructed images will tend to converge, resulting in a small ($\sigma_{ij}^2\ll1$) variance and hence an increasingly negative $S$. The use of a large sample set in this method reduces the sensitivity to fluctuations, whereas the method described above using \eqref{eq:KL}, while providing a more interpretable representation of information flow, can by construction only compare a single reconstruction image with the original input. In the results section below, we use both the relative entropy \eqref{eq:KL} and the
differential entropy \eqref{eq:diffent} within the reconstruction approach. While the differential entropy shows greater numerical stability,  we emphasize that the results of both methods for computing the entropy associated with reconstructed images yields qualitatively and quantitatively similar results for the cutoff $\ell^*$.

A further advantage of the reconstruction method is that the encoding of the information in the hidden layers, after reconstruction, is similar to the encoding of the information in the input layer. This renders the information stored in the hidden layers accessible to the observer and provides an approach to understanding the decision-making process of neural networks. An example for this is given in fig.~\ref{fig:cascade-example}, where the digit seven can be identified in all reconstructions. A more involved example is given in fig.~\ref{fig:xai_channel_reconstruction} using a colored figure of two animals and the vgg16 convolutional neural network \cite{simonyan2015deepconvolutionalnetworkslargescale}. It is possible to identify certain features of the reconstructed images as particularly relevant for the network's decision-making. This may be of particular use when using machine learning in daily-life situations, where it is crucial to be able to understand the decisions taken by the network and to verify their suitability for the given task.

\newpage 

\section{Results}\label{sec:results}

Here we apply our reconstruction entropy  method to both MNIST \cite{deng2012mnist} and CIFAR10 \cite{CIFAR10}. Specifically, we compute the cutoff $\ell^*$ within the reconstruction approach for untrained networks of varying depths over a range of locations in the phase space parametrized by $(\sigma_w^2,\sigma_b^2)$. This gives us proxy measurement for the correlation length, and in particular a prediction for the location of the critical point. To test the accuracy of our predictions, we also train the networks, and compare the maximum trainable depth as measured by accuracy to the cutoff $\ell^*$. The networks are initialized and trained as in \cite{schoenholz2017deep}, but using a fixed learning rate of 0.001, a depth of $L=100$, and a maximum of 200 epochs of training. The corresponding code for this work is freely available on GitHub \cite{GITHUB}. 

\subsection{Detecting trainability prior to training}

As described in the Introduction, we determine the trainability of the networks prior to training by computing the reconstruction entropy. Intuitively, this quantifies the depth to which information about the input can propagate, thus providing us with a measure of the correlation length (see \cite{schoenholz2017deep} for a derivation of the correlation length in the infinite-width limit; in contrast, our method does not require large width). Due to numerical fluctuations, we must choose a threshold $\eta$ in order to determine the saturation point (see Methods below). We consider both the relative and the differential entropy within the reconstruction approach.

\begin{figure}[H]
    \centering
\includegraphics[width=0.8\textwidth]{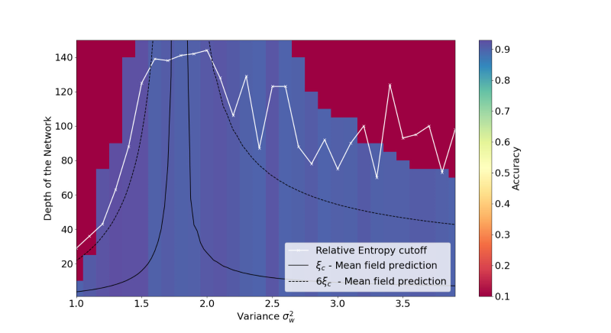}
    \caption{ \textbf{Predicting trainability with reconstruction relative entropy}. The figure shows trainability on MNIST along a fixed $\sigma_b^2=0.05$ slice through the 2d phase space $(\sigma_w^2,\sigma_b^2$), cf. \cite{schoenholz2017deep}. The colour corresponds to the final accuracy obtained by the networks after 200 epochs of training. Overlaid  is the  reconstruction relative entropy cutoff $\ell^*$ for a threshold value of $\eta=0.005$ (white curve), as well as the $N\to\infty$ central limit theorem prediction (black curves). The relative entropy cutoff accurately predicts the location of the critical point, providing a proxy for the correlation length in the system.  We emphasize that the reconstruction entropy curves are obtained prior to training the DNN, i.e., for \emph{untrained} networks, and are thus orders of magnitude cheaper to generate. 
    \label{fig:combination1}}
\end{figure}

For the relative entropy, fig.~\ref{fig:combination1} shows the results for the cutoff $\ell^*$ and the accuracy of neural networks for a fixed-width ($N_{\ell<L-1}=784$, $N_{L-1}=400$) network for MNIST. As shown in  fig.~\ref{fig:combination1}, the cutoff $\ell^*$ - shown in white - qualitatively predicts the empirical correlation length as measured by the accuracy of the trained network. The exact cutoff value depends on the threshold for numerical fluctuations (see Methods below); we have found that a threshold value of around $\eta=0.005$ provides the best quantitative match, though the ordered phase (low $\sigma_w^2$) appears less sensitive to this choice than the chaotic phase (high $\sigma_w^2$). This is unsurprising in light of the size of thermal fluctuations in either phase, cf. panel (f) of fig.~\ref{fig:cutoff_detection}. Notably, the location of the critical point -- determined in the reconstruction entropy method as the point at which $\ell^*=L$ -- is relatively robust to the choice of threshold value, and exhibits excellent agreement with the critical regime determined by actually training the DNN.

\begin{figure}[h!]
    \centering  
    \includegraphics[width=\linewidth]{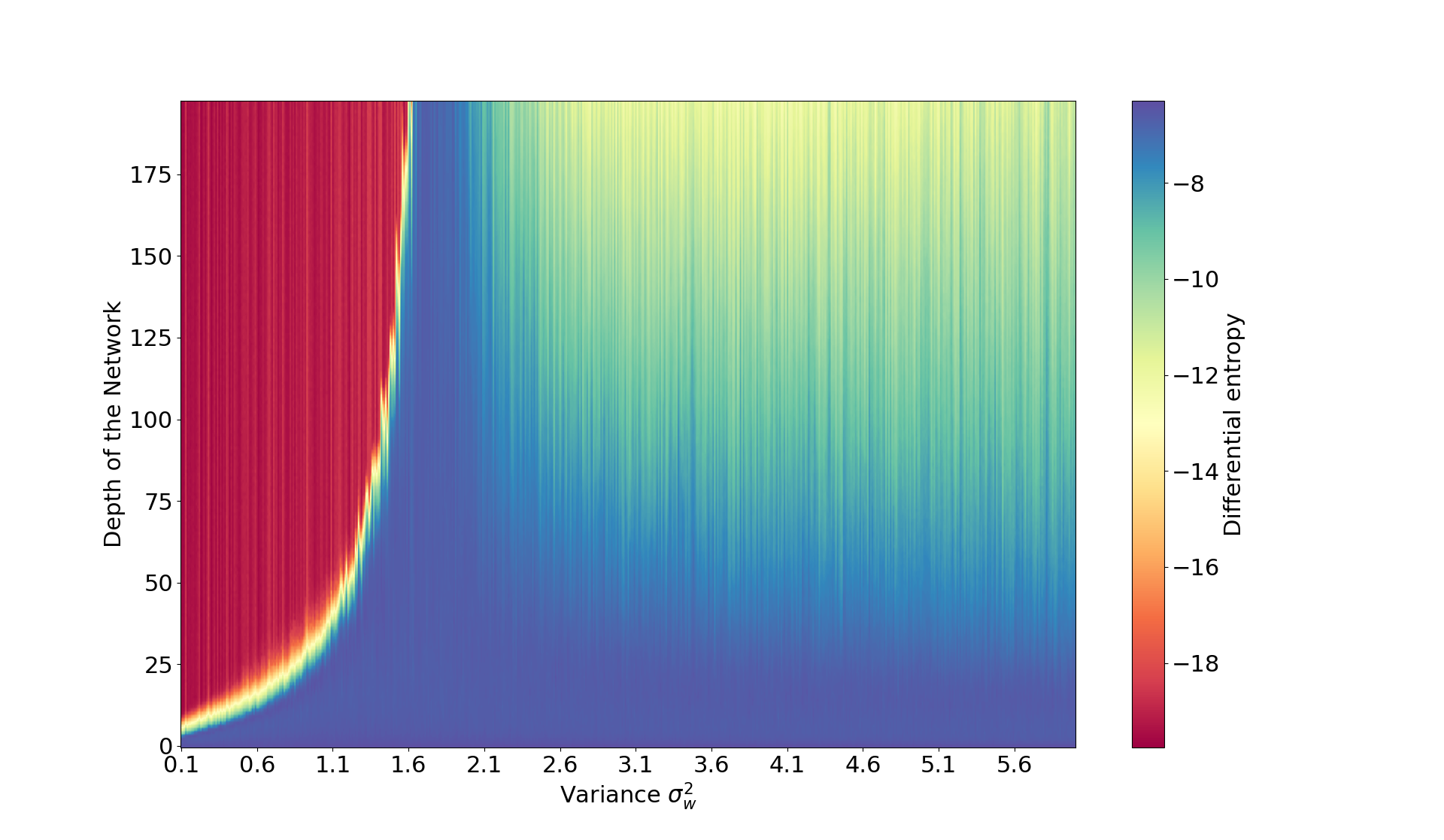}
    \caption{ \textbf{Predicting trainability with reconstruction differential entropy}. The figure shows the measurement of differential entropy given by \eqref{eq:diffent2} for a wide neural network 200 layers deep, trained on MNIST. The entropy is calculated via the reconstruction method prior to any training. As shown in the color scheme to the right, red corresponds to large information loss (i.e., untrainable), while blue corresponds to more information being preserved (i.e., more trainable), as quantified by the differential entropy.
    The ordered phase is clearly visible (left side) due to the rapid decay of correlations. This rapid information loss causes a clear delineation between the trainable and untrainable regimes.
    In the chaotic regime (right side), information is lost much more gradually, such that sufficiently long training will eventually allow deeper networks to train successfully, albeit with potentially significant costs (c.f. \cite[fig.~7]{schoenholz2017deep}).
    Around the critical point of $\sigma_w^2 = 1.76$, information is preserved even for deeper layers (blue cone). These predictions, obtained prior to training, are in agreement with the empirical observations for trained networks shown in \cite{schoenholz2017deep}.
    \label{fig:diff_entropy}}
\end{figure}

\begin{figure}[h!]
    \centering  
    \includegraphics[width=\linewidth]{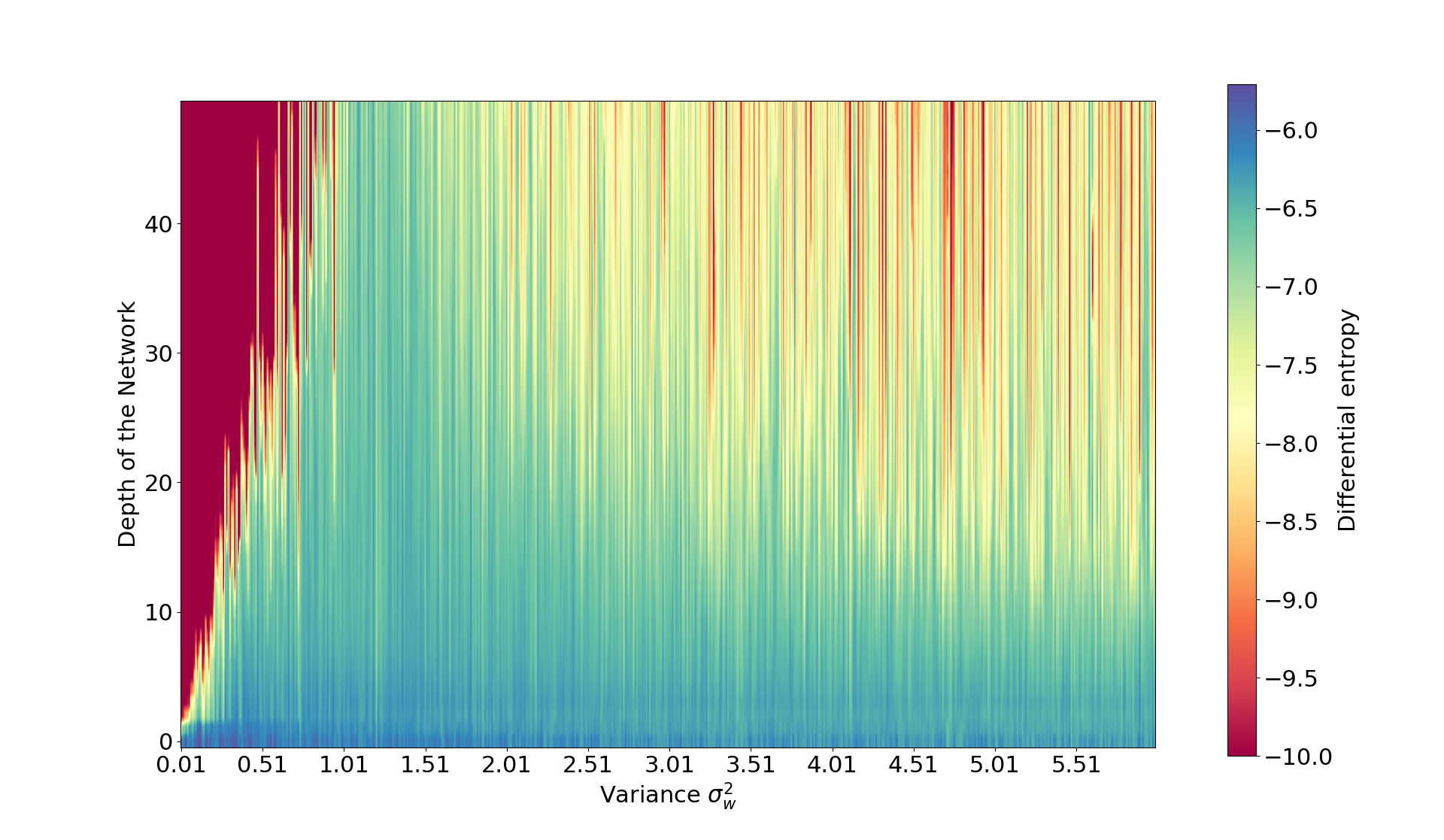}
    \caption{ \textbf{Predicting trainability for a ResNet with reconstruction differential entropy}. The figure shows the predicted trainability with reconstruction differential entropy for a ResNet trained on FashionMNIST. The bias variance is set to $\sigma_b^2=5e^{-2}$. 
    We observe a similar phase behaviour as in fig. \ref{fig:diff_entropy}, though the edges of the trainable region are less sharp. Particularly in the chaotic phase (towards the right), the large fluctuations seem to indicate the role of residual connections in preserving relevant information despite the otherwise limited signal propagation. Nonetheless, one clearly sees an optimum critical region at approximately $\sigma_w^2\sim 1.3$ \label{fig:diff_entropy_resnet}}
\end{figure}

In fig.~\ref{fig:diff_entropy}, the differential entropy is shown for a wide neural network with 200 layers as function of the weight variance. The network is untrained and the differential entropy is calculated for the MNIST data. 
The numerical stability of the differential entropy is indeed significantly improved as compared to the relative entropy shown in fig.~\ref{fig:combination1}.
In fig.~\ref{fig:diff_entropy}, the three different phases can easily be identified: On the left hand side, the network is in the ordered phase and information is lost within a few ($O(1)$) layers. This is reflected in a large negative differential entropy. At the critical point, the differential entropy remains almost unchanged over the depth of the network. In the chaotic regime, the differential entropy decreases with increasing depth. However, the information loss is slower than in the ordered regime. This can be seen as the differential entropy in the ordered regime is around $-18$ for the deeper networks, whereas for the chaotic phase similar deep networks cause a differential entropy of around $-13$.  Moreover, the behaviour is less stable and more fluctuations are observed. The image shown  again highlights the relation between information content and trainability: close to criticality the information loss over depth is limited and the network is thus able to learn. For the other regions the information is lost at some depth, reducing the trainability. As the evaluation of trainability depends on a number of maximal epochs to train, the exact information content for training a network is unknown. However, the more information reaches a given depth of the network, the faster the network will be able to learn and the higher its accuracy will be. Figure \ref{fig:diff_entropy_resnet}, shows similar results for a ResNet trained on FashionMNIST. The bias variance is set to $\sigma_b^2=2e^{-5}$, similar to the CNN in \cite{xiao2018dynamical} as its residual blocks consist of convolution layers. The ResNet consists of 50 residual blocks. Each block consists of 2 convolutional layers (cf.~\cite{he2015deepresiduallearningimage}). Ignoring the skip connections, the ResNets are thus similar to 100 layer deep convolutional networks.

\subsection{Reduction in overall training time}

The reconstruction entropy  method results in a significant overall reduction in training time for two reasons: first, a single epoch of training on the shallow reconstruction networks that comprise the cascade is significantly faster than an epoch of training on the deep feedforward network, and is all that suffices for the reconstruction networks to reach sufficiently high accuracy. Second, the ability to determine the trainable regime in phase space without needing to actually train the DNN results in a decoupling between parameters and hyperparameters, thereby significantly reducing the time complexity of computationally costly grid searches over combinations thereof. Here we describe these cost savings in more detail.
\begin{table}[h!]
    \centering
    \begin{tabular}{c|c||c|c|c}
        Dataset & Type & Single epoch [s] & Prediction [s] & Single reconstruction [s]\\
        \hline \hline
        MNIST & Wide & $(10.357 \pm 0.051)$ & $(28.94 \pm 0.34)$ & $(0.1513 \pm 0.0038)$\\
        MNIST & Shrinking & $(9.106 \pm 0.063)$ & $(29.43 \pm 0.20)$ & $(0.1525 \pm 0.0042)$\\
        \hline
        CIFAR10 & Wide & $(11.425 \pm 0.052)$ & $(24.80 \pm 0.25)$ & $(0.1262 \pm 0.0034)$\\
        CIFAR10 & Shrinking & $(9.054 \pm 0.052)$ & $(25.31 \pm 0.22)$ & $(0.1271 \pm 0.0034)$\\
    \end{tabular}
    \caption{\textbf{Benchmarks performed using a Intel Xeon w3-2435 CPU and NVIDIA RTX A5500 GPU.} Two networks with either constant layer size (Wide) or a linear shrinking layer size (Shrinking) are tested for both datasets. The corresponding type is denoted in the second column. The initial layer size of the networks is fixed by the training data. In the third column, the times for training such a deep network one single epoch are shown. The shown times are average values over 100 runs and their sample standard deviation as uncertainty. In the fourth column, the average time for predicting the reconstruction cutoff in these networks is shown. We stress that this implementation of the prediction is not optimized, and does \emph{not} use parallelization; it should thus be seen as an upper bound of the true required time (e.g., if reconstruction networks are trained in parallel). For this reason, in the last column we also give the time for training a single reconstruction network. Theoretically, all the reconstruction networks can be trained in parallel which should reduce the prediction time in the fourth column significantly. We note that typically, many tens or hundreds of training epochs are required for training the DNN itself, so the third column is an extreme lower bound. As discussed in the main text, for a particular example we find a speed-up of $\mathcal{O} (100)$.
    \label{tab:benchmarks}}
\end{table}

The training time for the reconstruction networks (i.e., the layers of the cascades) for a DNNs is benchmarked in table \ref{tab:benchmarks}.
The time to compute the differential entropy \eqref{eq:diffent} is around $0.02 s$ and thus negligible compared to the training time of the reconstruction networks. In our tests, we observe that predicting the information propagation depth takes roughly as long as training a network for three epochs. 
In practice, the number of epochs to observe trainability is dependent on other hyperparameters including batch size and learning rate. Inappropriate hyperparameters may result in an non-learning behaviour even close to criticality. Let us give an example for speedup through our method, as compared to a typical number of epochs of training, assuming sufficiently appropriate chosen hyperparameters besides $\sigma_w^2$. As example we recall \cite[fig.~7]{schoenholz2017deep}, where the critical region becomes visible after 300 epochs of training. For this case, our method provides a speedup of a factor of $O(100)$.

In practice, the choice of hyperparameters such as the learning rate also play an important role in trainability (hence the widespread use of optimizers such as ADAM \cite{ADAM}). In general, the brute-force method for determine the optimum initial conditions is to perform a computationally costly grid search over all possible combinations. One example of optimization algorithms is optuna\footnote{See \cite{optuna_2019} and https://optuna.readthedocs.io/en/stable/}. The associated time complexity is a product of the number of values of each (hyper)parameter. As a simplified example,  if we limit ourselves to $M_w$ values of the weight variance $\sigma_w^2$ and $M_r$ values of the learning rate $r$, then the time complexity for this search is $O(M_w M_r)$. From the phase space perspective taken here however, it is more efficient to first determine the trainable regime in parameter space, and only then select the optimum hyperparameters, cf. footnote \ref{ft:hyper}. This reduces the time complexity to $O(M_w+M_r)$.\footnote{More generally, we have $O\left(M_wM_b\prod_iM_i\right)\to O\left(M_wM_b+\prod_iM_i\right)$ where $M_b$ is the number of bias variances and $M_i$ the number of the $i^\mathrm{th}$ hyperparameter over which we scan.} In other words, we split the grid search into a search over phase space $(\sigma_w^2,\sigma_b^2)$ followed by a search over hyperparameters. At each point in the former, we enjoy the speedup provided by the RE method for determining whether a network at that point is trainable for some choice of the latter.\footnote{That is, initialization in the critical regime (more generally, $L\leq\ell^*$) is a necessary but not sufficient condition for trainability in the sense that one must still select hyperparameters, but -- assuming one is below the line of uniformity identified in \cite{Bukva:2023ksv} -- it is a necessary and sufficient condition in the sense that \emph{some} choice of hyperparameters will work.} In fact, we expect the time complexity of locating the critical value of $\sigma_w^2$ itself to be nearer $O(\ln M_w)$, since the form of the cutoff curves in fig.~\ref{fig:combination1} is amenable to a binary search, modulo fluctuations.

\subsection{Opening the black box: retracing the neural network's decisions}

\begin{figure}[h!]
    \centering
    \includegraphics[width=0.9\textwidth]{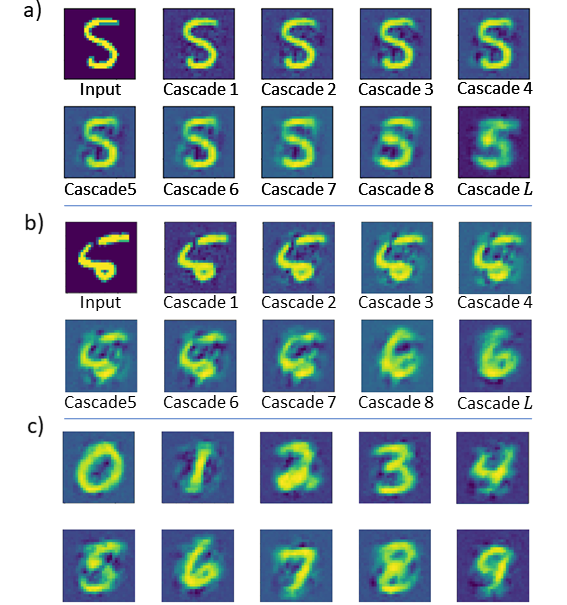}
    \caption{\textbf{Visualizations of the network's decision process. } (a) Reconstruction of the sequence of layers of a shrinking neural network with 10 hidden layers, initialized with $\sigma_w = 1.65$ and trained for 20 epochs. The variance of $\sigma_w = 1.65$ is chosen  close to the critical point and thus in the well-trainable regime. The input is drawn from the MNIST dataset. The input image is labeled as a 5. As can be seen, the features of the digit change over the depth of the network. Most prominently, the left upper edge at the horizontal bar of the 5 becomes more sharp. (b) Same as (a), except that here the network predicts a 6, which is shown in the lower row of the panel. The network adds a smooth connection between two sections of the 5, causing the internal representation to resemble a 6. (c) Reconstruction of the optimal digits as represented by the network.
    \label{fig:optimal_digits}}
\end{figure}

A further advantage of our reconstruction method is the possibility of retracing the network's decision-making process and obtaining insight into what the network considers at each depth. This is important  as neural networks are largely considered black boxes, and the reasons for their decisions are, in most cases, unknown. Other methods such as layer-wise relevance propagation \cite{1910.09840} allow one to interpret the importance of the input features, but do not provide insight into the representation of these features in hidden layers. Our reconstruction method provides visual insight into the decision-making process, allowing one to observe which features of the input image are preserved or altered, and to what extent, at a qualitative level. This handling of the data, e.g., preserving, enhancing, sharpening, or removing features, is highly dependent on the corresponding input. As an example consider fig.~\ref{fig:optimal_digits}, which corresponds to a wide network with $\sigma_w^2 = 1.6$ and $\sigma_b^2 = 0.05$, trained for 20 epochs on MNIST. The network has a depth of 10 hidden layers, and reaches an accuracy of $96\%$. In panel (a), the network correctly predicts the label of the input image as five. Over the reconstructions for the different depths, a change in features is observed. The most prominent change is the sharpening of the upper left edge of the digit shown, connecting the horizontal bar and the body of the image of the five. This is an example of the network enhancing a certain feature in the input data, indicating the importance of the considered feature for the classification. In contrast, in panel (b), the image of a five is incorrectly classified  as a six. In the first layers, the network closes the distance between the horizontal bar on top of the five and its main body. At a depth of six layers, the main body and the horizontal bar are connected. In the subsequent layers, the network creates a smoother transition between the main body and the horizontal bar. This change is directly opposite to the enhancement of this specific feature in (a), and ultimately results in the misclassification. This highlights the different internal treatments of the two images (both of which correspond to the same digit), and sheds light on why the network gives different predictions for them. 

Moreover, we may use our method to reconstruct the images that the network considers perfect examples of the corresponding label, thereby gaining insight into how this information is represented internally. To this end, we create an output vector with maximal activation on the label chosen as an example. Fig. \ref{fig:optimal_digits}, panel (c), shows the optimal digits for the given network. \\

\begin{figure}
    \centering
    \includegraphics[width=0.75\linewidth]{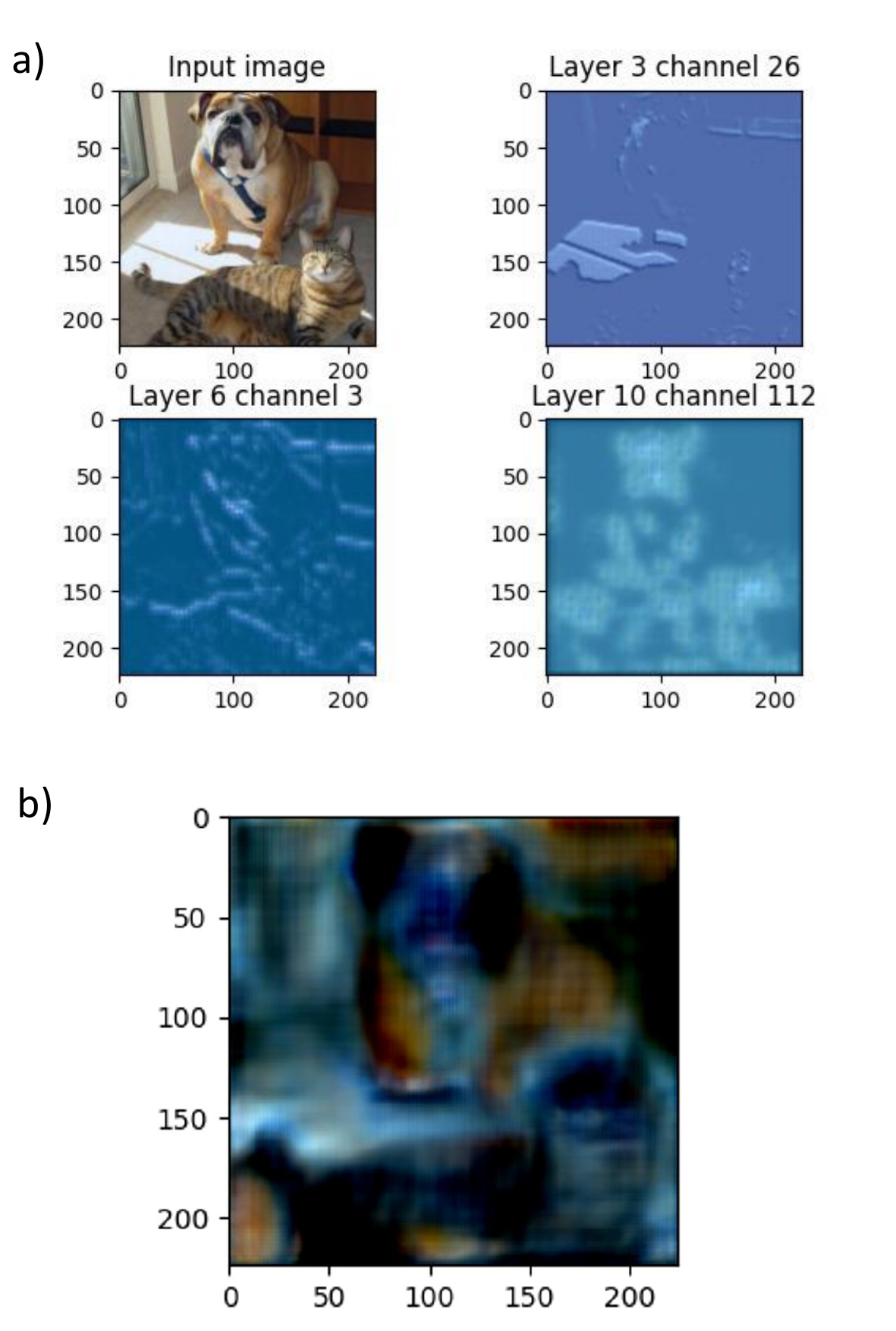}
    \caption{Panel (a) shows selected reconstructions of different channels of different layers for the input image and the vgg16 convolutional network \cite{simonyan2015deepconvolutionalnetworkslargescale}. The original image is taken from \cite{Selvaraju_2019}. For each layer we present a single example of channel, which shows a particular characteristic of the image (e.g., brightness and edges). Different channels in the layers yield different features of the image. Panel (b) shows a reconstruction using \textbf{all} channels, starting at the last pooling layer at a depth of 17.}
    \label{fig:xai_channel_reconstruction}
\end{figure}

As a more complex proof of concept, we apply the reconstructions to the convolutional network vgg16 \cite{simonyan2015deepconvolutionalnetworkslargescale}. The trained weights for this network are taken from the torch library \cite{torchlib}. We train the cascade networks by $30,000$ examples form the ImageNet dataset \cite{5206848} for a single epoch each. We use the test image of \cite{Selvaraju_2019}, displaying a cat and a dog in front of a window as an example.
Using these cascade networks, we are able to reconstruct the information propagating to any given layer, as well as the information within individual channels. We use the activations in all channels of a given layer of the CNN for the reconstruction (cf. fig. \ref{fig:xai_channel_reconstruction} b). Alternatively, we may suppress the information in different channels and focus on a single one instead (cf. fig. \ref{fig:xai_channel_reconstruction} a). The reconstruction method allows us to determine which properties -- e.g., degree of brightness, presence of edges or animals -- are stored in different channels. For example, in fig. \ref{fig:xai_channel_reconstruction} a), a selection of reconstructions of different channels in different layers of the CNN are shown: in the top right panel, titled ``layer 3 channel 26'' the brightness is memorized by the network. This includes the bright spot on the floor from the window and the left side of the dog. In the lower left panel, titled ``layer 6 channel 3'', the channel focuses on edges. In the lower right panel, titled ``layer 10 channel 112'', bright spots mostly belonging to the cat and dog are focused on. Moreover, in fig. \ref{fig:xai_channel_reconstruction} b), the full reconstruction, i.e. the reconstruction using all channels after all convolution and pooling layers, is shown. Obviously, the original image is now blurred. That is expected as vgg16 involves multiple Maxpooling layers, causing a loss of local information and the network in general is expected to decrease the information with increasing depth. In this case, vgg16's most probable prediction for the test image is the class with label 243, which predicts a ``bull mastiff'' in the image.

\section{Discussion}

Our approach uses single-layer reconstruction networks with the aim of minimizing differences in the representation of meaningful information from one hidden layer to the next. By computing the entropy between the reconstructed images and the original inputs, we thus obtain a measure of the information lost at subsequent layers. Importantly, the qualitative features of the resulting entropy curves are sensitive to the phase space of the network, and hence -- by indicating the critical surface -- can be used to predict trainability \emph{without} training the feedforward network. This stands in contrast to \cite{schoenholz2017deep} and related works, which required many hundreds of epochs of training; additionally, these previous works rely on the central limit theorem (and hence large width) to predict the critical regime with accuracy, whereas our reconstruction method works for arbitrary layer widths. This significantly reduces the computational cost of training deep networks, since the network can be initialized in optimal regions of phase space without needing to perform expensive grid searches. 

Furthermore, our reconstruction method yields qualitative insights into the representation of information in hidden layers, and can be used to identify key features that determine network behaviour (cf. fig.~\ref{fig:optimal_digits}). It also provides a reconstruction of the network's internal representation of each category; e.g., in panel (c) of fig.~\ref{fig:optimal_digits}, we see what the trained network expects each MNIST digit to look like.

We expect that our results will have a significant impact on  the practical costs of training large networks. Moreover, our method provides insight into how information is represented and used for decision-making processes, thereby taking us one step closer to Explainable AI.

An obvious technical improvement would be to train the reconstruction layers in parallel. The gains from parallelization are highly hardware-dependent and we have not explored this here. For networks with $O(10^2)$ layers, this could potentially lead to a further speedup of two  additional orders of magnitude. Additionally, we have demonstrated the effectiveness of this reconstruction method only for fully-connected feedforward networks (MLPs). A further task is to expand this method to more complex architectures, such as convolutional neural networks, recurrent neural networks, and transformers.


\begin{appendices}
	
\section{Methods}\label{sec:hist}

As described in the Introduction, we wish to compute the entropy between an image reconstructed from layer $\ell\in(0,L]$ and the original input, which we dub \emph{reconstruction entropy}. To do this, we train a shallow auxiliary network to reconstruct $\mathbf{z}^{\ell-1}$ from $\mathbf{z}^\ell$. A sequence of these single-layer reconstruction networks is then combined to form an $\ell$-layer \emph{cascade}, as shown in fig.~\ref{fig:cascade-example}, which produces a reconstructed image from the activations $\mathbf{z}^\ell$ of the feedforward DNN.

The essential idea is that if meaningful information in the DNN is perfectly preserved from layer $\ell$ to $\ell+1$, then we should find zero reconstruction entropy between the reconstructed image and the original input. In practice however, information is gradually lost with increasing depth, which causes the reconstructed image to degrade; see fig.~\ref{fig:combination1}. In both the ordered and chaotic phases, we observe an abrupt saturation of the relative entropy as all residual information is lost at some finite ($\ell^*<L$) layer, signifying that networks with a depth of $L>\ell^*$ are untrainable at this point in phase space.

\subsection{Quantitative description of the cascade method}
\label{sec:theoback}
To describe our method, let
\begin{equation}
	f^{\ell}: \mathbb{R}^{N_{\ell-1}} \rightarrow \mathbb{R}^{N_\ell}~,
	\quad\quad
	\ell\in[1,L]~,
\end{equation}
be the map from the preactivations $\mathbf{z}^{\ell-1}\in\mathbb{R}^{N_{\ell-1}}$ to the activations $\mathbf{z}^\ell\in\mathbb{R}^{N_\ell}$ given by a single layer of the feedforward network, cf. \eqref{eq:DNN}. An $L$-layer DNN can then be defined as the composition of these functions,
\begin{equation}
    F^L(\mathbf{x}) \coloneqq \left(f^L\circ ...  \circ f^1\right)(\mathbf{x})~,
\end{equation}
where $\mathbf{z}^0=\mathbf{x}$ is the vectorized input image (e.g., MNIST or CIFAR10). Our approach is to introduce an auxiliary reconstruction network for each $f^\ell$,
\begin{equation}
	c^\ell: \mathbb{R}^{N_\ell}\to\mathbb{R}^{N_{\ell-1}}~,
\end{equation}
which is trained to reconstruct the corresponding preactivations, i.e., 
\begin{equation}
	\left(c^\ell \circ f^\ell\right)(\mathbf{z}^{\ell-1}) = \mathbf{z}^{\ell-1}~.
    \label{eq:ci_fi_one}
\end{equation}
As written, this represents the idealized case of perfect reconstruction, i.e., ${c^\ell=\left(f^\ell\right)^{-1}}$. In practice, this equation will only be satisfied approximately, so that the reconstruction network is not simply the inverse of the corresponding feedforward layer; we will discuss this momentarily. For now, let us proceed with the idealized reconstruction \eqref{eq:ci_fi_one}. The composition of shallow reconstruction networks,
\begin{equation}
	\mathcal{C}^\ell\coloneqq\left(c^1\circ\ldots\circ c^\ell\right)~,
\end{equation}
is called a \emph{cascade}. The reconstruction of the input from a given layer $\ell$ is then obtained via
\begin{equation}
	\mathbf{x}=\left(\mathcal{C}^\ell\circ F^\ell\right)(\mathbf{x})
	=\left(c^1\circ\ldots\circ c^\ell\right)\circ\left(f^\ell\circ ...  \circ f^1\right)(\mathbf{x})~.
	\label{eq:totalrecon}
\end{equation}
Thus, we obtain reconstructions at successively deeper layers of the DNN by evaluating \eqref{eq:totalrecon} at each $\ell\in[1,L]$.

In practice however, the reconstruction process will not be perfect, and each reconstruction network will be associated with some error $\boldsymbol{\epsilon}_i^\ell$, so that in place of \eqref{eq:ci_fi_one} we have
\begin{equation}
	\left(c^\ell \circ f^\ell\right)(\mathbf{z}^{\ell-1}) = \mathbf{z}^{\ell-1}+\boldsymbol{\epsilon}^\ell~.
\end{equation}
Thus, in practice, the total reconstruction \eqref{eq:totalrecon} becomes
\begin{equation}
	\left(\mathcal{C}^\ell\circ F^\ell\right)(\mathbf{x})
	=\mathbf{x}+\sum_{m=1}^\ell \boldsymbol{\epsilon}^m\prod_{n=0}^{m-1}\partial c^n(\mathbf{z}^n)\eqqcolon\bx~,
	\label{eq:reconerr}
\end{equation}
to leading order in the reconstruction error $\epsilon^\ell\!\simeq\!\epsilon\!\ll\!1$. To obtain this expression, observe that
\begin{equation}
	\left(c^1\circ\ldots\circ c^\ell\right)\circ\left(f^\ell\circ ...  \circ f^1\right)(\mathbf{x})
	=\left(c^1\circ\ldots\circ c^\ell\right)(\mathbf{z}^\ell)~.
\end{equation}
Thus, the first cascade yields
\begin{equation}
	\left(\mathcal{C}^1\circ F^1\right)(\mathbf{x})=\left(c^1\circ f^1\right)(\mathbf{x})=c^1(\mathbf{z}^1)=\mathbf{x}+\boldsymbol{\epsilon}^1~.
\end{equation}
Working to linear order in the error, the next yields
\begin{equation}
	\begin{aligned}
		\left(\mathcal{C}^2\circ F^2\right)(\mathbf{x})&=\left(c^1\circ c^2\circ f^2\circ f^1\right)(\mathbf{x})=\left(c^1\circ c^2\right)(\mathbf{z}^2)\\
		&=c^1(\mathbf{z}^1+\boldsymbol{\epsilon}^2)=c^1(\mathbf{z}^1)+\boldsymbol{\epsilon}^2\partial c^1(\mathbf{z}^1)
		=\mathbf{x}+\boldsymbol{\epsilon}^1+\boldsymbol{\epsilon}^2\partial c^1(\mathbf{z}^1)~,
	\end{aligned}
\end{equation}
where the partial denotes the derivative of the function with respect to the argument. The third cascade yields
\begin{equation}
	\begin{aligned}
		\left(\mathcal{C}^3\circ F^3\right)(\mathbf{x})&=\left(c^1\circ c^2\circ c^3\circ f^3\circ f^2\circ f^1\right)(\mathbf{x})=\left(c^1\circ c^2\circ c^3\right)(\mathbf{z}^3)\\
		&=\left(c^1\circ c^2\right)(\mathbf{z}^2+\boldsymbol{\epsilon}^3)
		=c^1\left(c^2(\mathbf{z}^2)+\boldsymbol{\epsilon}^3\partial c^2(\mathbf{z}^2)\right)\\
		&=\left(c^1\circ c^2\right)(\mathbf{z}^2)+\boldsymbol{\epsilon}^3\partial c^2(\mathbf{z}^2)\,\partial\!\left(c^1\circ c^2\right)(\mathbf{z}^2)\\
		&=c^1(\mathbf{z}^1+\boldsymbol{\epsilon}^2)+\boldsymbol{\epsilon}^3\partial c^2(\mathbf{z}^2)\,\partial c^1(\mathbf{z}^1+\boldsymbol{\epsilon}^2)\\
		&=c^1(\mathbf{z}^1)+\boldsymbol{\epsilon}^2\partial c^1(\mathbf{z}^1)+\boldsymbol{\epsilon}^3\partial c^2(\mathbf{z}^2)\left[\partial c^1(\mathbf{z}^1)+\cancel{\boldsymbol{\epsilon}^2\partial^2 c^1(\mathbf{z}^1)}\right]\\
		&=\mathbf{x}+\boldsymbol{\epsilon}^1+\boldsymbol{\epsilon}^2\partial c^1(\mathbf{z}^1)+\boldsymbol{\epsilon}^3\partial c^2(\mathbf{z}^2)\,\partial c^1(\mathbf{z}^1)~,
	\end{aligned}
\end{equation}
where we have dropped terms second-order and higher in the error. Continuing in this manner yields \eqref{eq:reconerr}.

The Kullback-Leibler divergence is then calculated between $\mathbf{x}$ and $\bx$, interpreted as probability mass functions. That is, let the original inputs be normalized such that $\sum_ix_i=1$. Then in \eqref{eq:KL}, we may take $p(x_i)=x_i$, and $q(\bar x_i)=\bar x_i/\sum_i\bar x_i$. We then observe that to leading order in the errors,
\begin{equation}
	\ln\frac{\bar x_i}{\sum_i\bar x_i}
	\approx\ln x_i+\sum_{n\neq i}\sum_{m=1}^\ell \epsilon_n^m\prod_{n=0}^{m-1}\partial c^n(\mathbf{z}^n)
	-\frac{1}{x_i}\sum_{n\neq i}x_n\sum_{m=1}^\ell \epsilon_i^m\prod_{n=0}^{m-1}\partial c^n(\mathbf{z}^n)~.
\end{equation}
Therefore, the KL divergence \eqref{eq:KL} quantifying the reconstruction entropy is
\begin{equation}
	\begin{aligned}
		D\big(&p(x)||q(x)\big)
		=\sum_ix_i\ln x_i-\sum_ix_i\ln\frac{\bar x_i}{\sum_i\bar x_i}\\
		&\approx-\sum_ix_i\sum_{n\neq i}\sum_{m=1}^\ell \epsilon_n^m\prod_{n=0}^{m-1}\partial c^n(\mathbf{z}^n)
			+\sum_i\sum_{n\neq i}x_n\sum_{m=1}^\ell \epsilon_i^m\prod_{n=0}^{m-1}\partial c^n(\mathbf{z}^n)~,
	\end{aligned}
\end{equation}
which goes to zero in the idealized case of zero error, when \eqref{eq:ci_fi_one} is realized (i.e., $p=q$). Here we see that additional layers contribute additively to the divergence. Na\"ively, this would seem to imply a gradual increase in error rather than the abrupt degradation observed in \ref{fig:cutoff_detection}. 
However, in the regime where the relative entropy saturates, the errors will be large, and hence the series expansion based on the assumption that $\epsilon^\ell\ll1$ breaks down long before this behaviour is observed.

\subsection{Determining the cutoff}
\label{sec:num_methods}

As explained in the introduction, our central claim is that the cutoff $\ell^*$ at which the reconstruction entropy saturates is a proxy for the correlation length of the network. Here we discuss some details associated with determining this value.

In panel (f) of fig.~\ref{fig:cutoff_detection}, we display a representative plot of the KL divergence \eqref{eq:KL} as a function of depth at three different locations in phase space. (The original input is used as the reference probability mass function $p(x)$, while the reconstructed image at subsequent layers is taken as the target function $q(x)$.) The key observation is that in the chaotic phase, the reconstruction entropy does not saturate, and instead continues non-monotonically increasing throughout the entire network. In contrast, saturation is observed in both the ordered and chaotic phases at some finite $\ell^*<L$. 

However, the saturation is not perfect, since the entropy continues to exhibit small fluctuations about this asymptotic value over all remaining layers. Therefore, to determine the cutoff numerically, we must choose a tolerance or threshold $\eta$, and proceed as follows: we set $\ell^*=L$ by default, and store the associated reconstruction entropy (RE). We then sequentially decrease the layer index $\ell_i$ until the difference between the RE at layer $L$ and that at layer $\ell_i$ exceeds the threshold $\eta$. At this point, we set $\ell^*=\ell_i+1$ (so that if the threshold is immediately exceeded, we recover $\ell^*=L$). As described in the main text, we have found that a value of $\eta=0.005$ works well in practice. If the value chosen for $\eta$ is too small, it will be immediately exceeded and one will have $\eta^*=L$; conversely, if the value is too large, the fluctuations will dominate over the curve. Thus, choosing the optimum value of $\eta$ is tantamount to selecting a suitable signal-to-noise ratio for the experiment. 

In addition to numerical fluctuations in the network however, computing RE using the KL divergence suffers from an additional systemic fluctuation due to the fact that, as mentioned in the Introduction, the reconstruction image to which all inputs converge after all information is lost may closely resemble one or more classes of inputs. This can be seen in fig.~\ref{fig:cutoff_detection}, where the similarity between the final reconstructed image in in panel (b) (equivalently panel (c)) and the digit 0 will be significantly smaller than that between this image and a 1. This makes the method using relative entropy prone to instabilities, i.e., we occasionally obtain spuriously large or small values of the cutoff for some inputs. To reduce the sensitivity to particular data samples, we average the relative entropy of an entire batch before determining the cutoff. Nonetheless, this method still exhibits fluctuations, as can be observed in the figure. 

A more stable approach is to instead compute RE using the differential entropy \eqref{eq:diffent}. As illustrated in fig.~\ref{fig:cutoff_detection}, this circumvents the problem because we compare different reconstructed images rather than reconstructed images and original inputs. The basic conceptual idea is however the same. In this case, once all information is lost, the set of reconstructed images for a diverse set of inputs will all be the same, resulting in vanishingly small variances for the probability distribution functions of the pixels. Thus, in the idealized limit where the reconstructions are exactly identical, the differential entropy \eqref{eq:diffent} will diverge to negative infinity (in practice, we find some large but finite value). Prior to information loss however, the reconstructed images will differ substantially (cf. panel (a) vs. (b) in fig.~\ref{fig:cutoff_detection}), and hence the differential entropy prior to the cutoff will be small. Unfortunately, the differential entropy is actually \emph{too} sensitive to remaining information, perhaps in part because it does not consider the original distribution of the input. This is not a problem in the stable regime, were we still observe a sharp prediction for the correlation length, cf. fig.~\ref{fig:diff_entropy}
. In the chaotic regime however, clear saturation is only observed for extremely large values of the variance $\sigma_w^2$, due to small but finite residual information in this phase.  Nonetheless, as is clear in fig.~\ref{fig:diff_entropy}, we still observe a region of high trainability (blue shading) consistent with previous results. In fact, this method may explain why the predicted correlation length in \cite{schoenholz2017deep} and related works in the chaotic phase is highly sensitive to the number of epochs, as (excessively) long training times will eventually allow the network to capitalize on this residual information, cf. the gradual shading in the chaotic phase in fig.~\ref{fig:diff_entropy}. In any case, the critical regime is clearly apparent, and again consistent with previous works, and in this sense the sensitivity of this method does not pose a problem in practice. We emphasize again that our prediction for this is obtained without training the feedforward network.

\subsection{Entropy of white noise}

\begin{figure}[h!]
    \centering  
    \includegraphics[width=\linewidth]{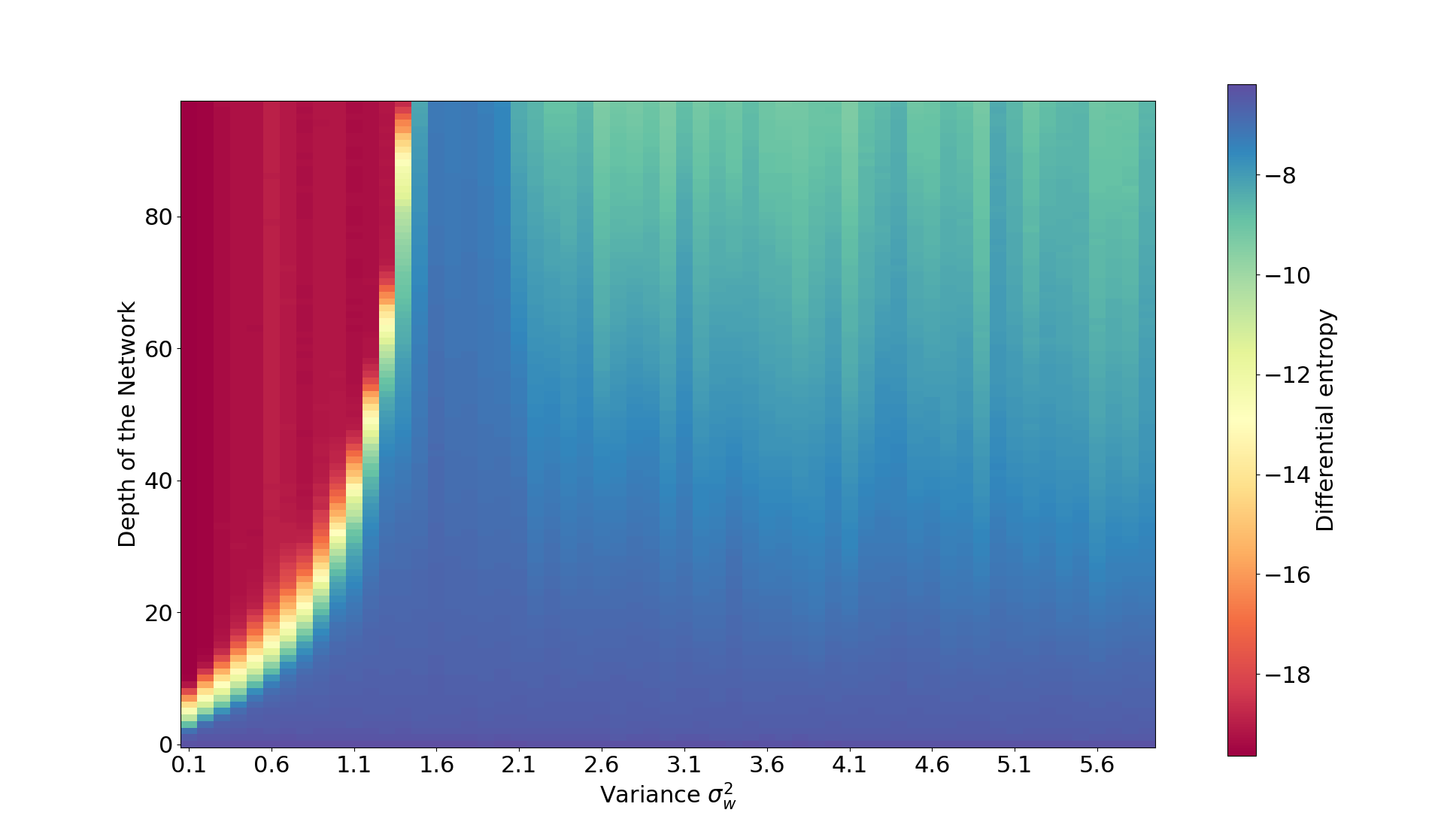}
    \caption{ \textbf{Predicting trainability for a wide neural network fed white noise with reconstruction differential entropy}. Here we show the same network architecture as in fig. \ref{fig:diff_entropy}, but instead of taking a structured dataset like MNIST, we feed the network white noise (again, no training of the feedforward network is performed). Despite this, the reconstruction differential entropy computed via the cascade layers is still able to distinguish the trainable regime.\label{fig:diff_entropy_noise}}
\end{figure}

To further illustrate that the information propagation is primarily a question of the location of the network in phase space rather than a property of the dataset itself, in fig. \ref{fig:diff_entropy_noise} we show the predicted trainability with reconstruction differential entropy for a DNN fed white noise. Again, we do not need to train the feedforward network: the point is that the cascades can still be trained to reconstruct the input at each layer, and that the associated reconstruction entropy computed with our method shows excellent agreement with fig. \ref{fig:diff_entropy}, in which the dataset was MNIST.

\subsection{Entropy of a CNN}
\begin{figure}[h!]
    \centering  
    \includegraphics[width=\linewidth]{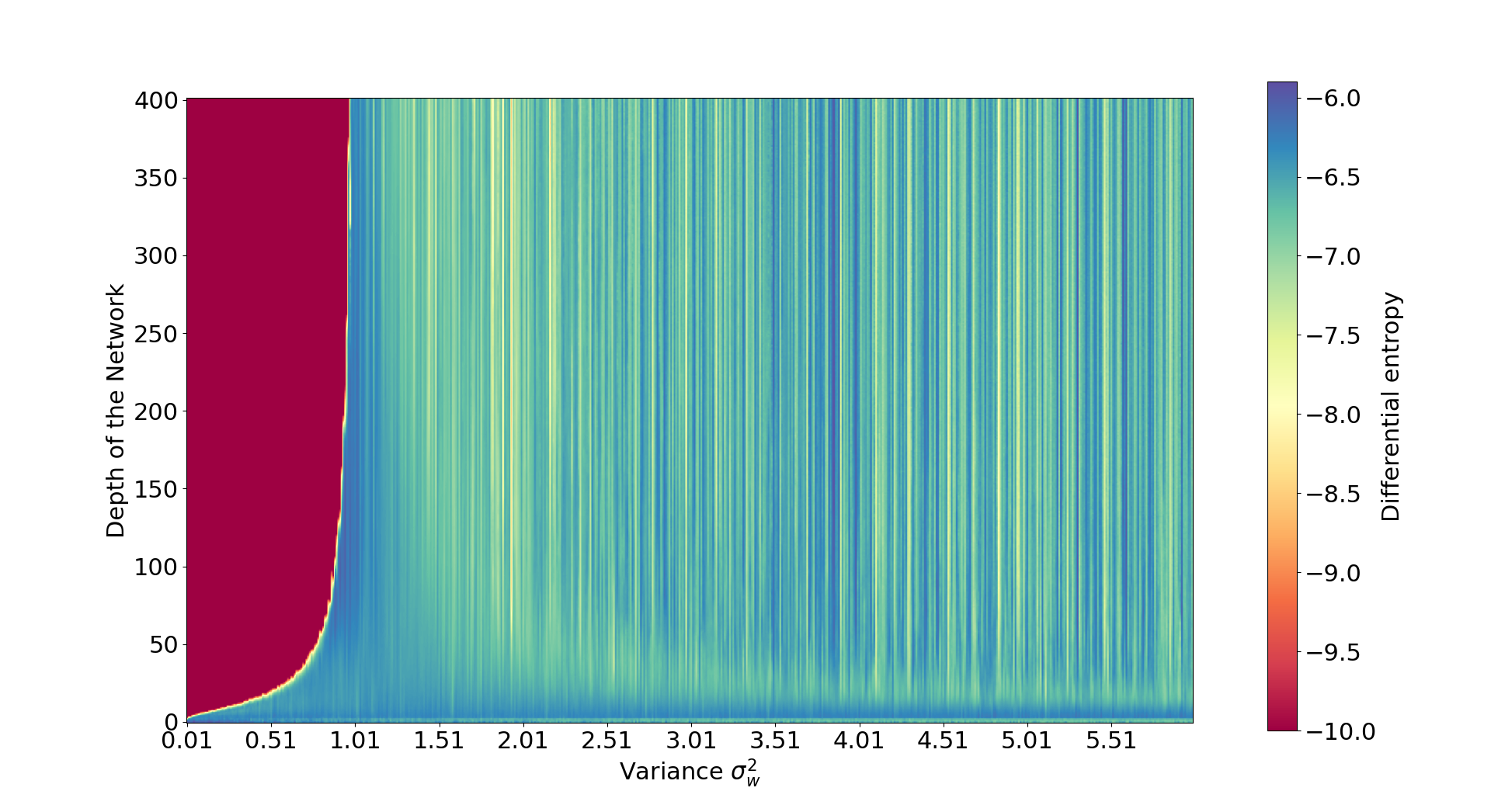}
    \caption{ \textbf{Predicting trainability for a CNN with reconstruction differential entropy}. Again we observe similar phase behaviour as in fig. \ref{fig:diff_entropy}, with a sharp delineation at the ordered phase boundary (left side), and a broad region of sub-optimal trainability in the chaotic phase right side). The optimum critical region is still visible as the densest peak near the ordered phase boundary.
    \label{fig:diff_entropy_cnn}}
\end{figure}

Fig.~\ref{fig:diff_entropy_cnn} shows the predicted trainability with differential entropy for a CNN trained on MNIST. For a better visualization the values for the differential entropy in the ordered phase are clipped at -10 to increase contrast.

\subsection{Pixel activations over images}\label{sec:toldyouso}

\begin{figure}
    \centering
    \includegraphics[width=0.8\textwidth]{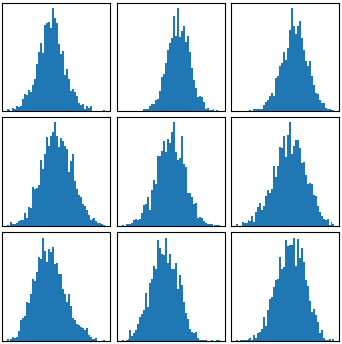}
    \caption{Histograms for the values of pixels 0 (left column), 391 (middle column), and 783 (right column) at layer 1 (top row), 10 (middle row), and 99 (bottom row) for a network initialized with variance $\sigma_w^2 = 3.4$. This corresponds to a network in the chaotic phase, but these plots are qualitatively unchanged at other points in phase space (e.g., in the ordered or critical regime). \label{fig:histograms}}
\end{figure}

The plots in Fig.~\ref{fig:histograms} substantiate the theoretical argument given above \eqref{eq:diffent2} that after reconstruction, the distribution of the activation of given pixel across images is approximately Gaussian. For a given pixel, we plot the distribution of the activation over a random batch of MNIST images after reconstruction. The histograms are shown for various layers in a DNN initalized at an arbitrary point in phase space. 

\end{appendices}

\bigskip
\clearpage
{\noindent \bf Acknowledgments.} We are grateful to Haye Hinrichsen, Luca Kohlhepp, Dominik Neuenfeld and Martin Rackl for useful discussions. JE and YT acknowledge support by the Deutsche Forschungsgemeinschaft (DFG, German Research Foundation) under Germany's Excellence Strategy through the W\"urzburg-Dresden Cluster of Excellence on Complexity and Topology in Quantum Matter - ct.qmat (EXC 2147, project-id 390858490). The work of JE and YT is also supported by DFG grant ER 301/8-1.

\noindent {\bf Author contributions.} The project was jointly initiated by all three authors. YT performed all coding involved. RJ and JE provided supervision and guidance. All three authors contributed to writing the manuscript. 

\noindent {\bf Corresponding author.} erdmenger@physik.uni-wuerzburg.de

\noindent {\bf Code and data availability and correspondence.} The code used for this work is publicly available (see \cite{GITHUB}).  Please address correspondence to erdmenger@physik.uni-wuerzburg.de.

\bibliographystyle{ytphys}
\bibliography{biblio}

\end{document}